%% file: main.tex
\documentclass[10pt,twocolumn,letterpaper]{article}

\usepackage[pagenumbers]{cvpr} 

\input{preamble}

\definecolor{cvprblue}{rgb}{0.21,0.49,0.74}
\usepackage[pagebackref,breaklinks,colorlinks,allcolors=cvprblue]{hyperref}

\def\paperID{334} 
\def\confName{3DV\xspace}
\def\confYear{2027\xspace}

\makeatletter
\newcommand*\bigcdot{\mathpalette\bigcdot@{2}}
\newcommand*\bigcdot@[2]{\mathbin{\vcenter{\hbox{\scalebox{#2}{$\m@th#1\bullet$}}}}}
\makeatother

\title{Differentiable Task-Driven Coded Aperture Camera \\Design with 3D Gaussian Scene Representations}

\author{Chengyang Yan \qquad Donald G. Dansereau\\
Australian Centre for Robotics, School of Aerospace, Mechanical and Mechatronic Engineering, \\ The University of Sydney\\
{\tt\small {chengyang.yan, donald.dansereau}@sydney.edu.au}
}

\begin{document}
\maketitle
\input{sec/0_abstract}    
\input{sec/1_intro}
\input{sec/2_lit}
\input{sec/3_method}
\input{sec/4_exp}
\input{sec/5_conclusion}

\newpage
{
    \small
    \bibliographystyle{ieeenat_fullname}
    \bibliography{main}
}

\clearpage \appendix \input{supplementary}

\end{document}

%% file: preamble.tex
\usepackage{multirow}
\usepackage{array}



%% file: sec/0_abstract.tex
\begin{abstract}
Camera design heavily influences perception task performance, yet cameras are typically designed by domain experts through costly experimentation, in isolation from the tasks they serve. Existing joint-design methods either rely on differentiable simulation limited to pre-captured 2D images, restricting them to optics or image-processing pipeline design, or use non-differentiable camera simulators optimised with derivative-free or reinforcement-learning methods that scale poorly with the number of design parameters. We present a differentiable framework for jointly designing complex camera systems and downstream perception tasks. We use it to design two coded-aperture depth-from-defocus cameras, a monocular and a stereo configuration, by jointly optimising the coded aperture mask, camera parameters, and a depth estimation network end-to-end. Our framework represents scenes with 3D Gaussian Splatting reconstructed from real-world captures, providing a differentiable image formation process for gradient-based optimisation over design spaces intractable for prior differentiable approaches. Results demonstrate that our differentiable method improves over existing coded-aperture design and non-differentiable camera design baselines, offering a step forward in automated and principled camera design for autonomous systems.
\end{abstract}

%% file: sec/1_intro.tex
\section{Introduction}
\label{diff-sec:intro}
\begin{figure}[ht!]
      \centering
      \includegraphics[width=\linewidth]{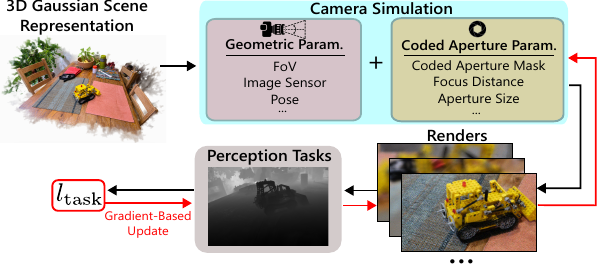}
      \caption{We introduce a differentiable method that jointly optimises camera parameters and perception tasks, using 3DGS scenes reconstructed from real-world captures and a coded aperture simulation modelling depth-dependent optical blur. Camera parameters and the perception network are jointly updated via gradient-based optimisation of the task loss (red arrows), spanning geometric parameters (FoV, pixel count, multi-camera configurations) and high-dimensional optical parameters (coded aperture mask).}
      \label{diff-fig:key_figure}
\end{figure}

The quality of camera captures directly governs the performance of downstream perception tasks, yet different autonomous systems require different camera designs to suit their tasks: autonomous vehicles need wide field of view (FoV) and high dynamic range, medical imaging needs fine resolution and colour fidelity, and robotics needs reliable depth and colour sensing. Hence, cameras should be designed in a task-specific manner.

Many works jointly optimise imaging systems and perception tasks via differentiable simulation and gradient-based optimisers~\cite{sitzmann2018end, chang2018hybrid, tseng2021differentiable, ikoma2021depth, cote2023differentiable, yang2023image, ghanekar2024passive}. However, their reliance on pre-captured 2D image datasets prevents generalisation to broader design problems such as FOV, resolution, multi-camera configurations, or unconventional architectures, limiting them to designing parts of the camera system (e.g., optics or the image signal processing pipeline) rather than full camera systems.

Other works design full camera systems using 3D simulation environments, but treat the simulator as a non-differentiable black box, relying on reinforcement-learning~\cite{klinghoffer2023diser} or derivative-free~\cite{yan2025tacos, tiwary2025if, yan2026joca} optimisers that ignore gradient information and must optimise for long iterations. This becomes prohibitively inefficient as the number of design parameters grows, particularly for high-dimensional parameters such as a coded aperture mask.

In this work, we propose a differentiable method for jointly designing complex camera systems and downstream perception tasks (Figure~\ref{diff-fig:key_figure}), using 3D Gaussian Splatting (3DGS)~\cite{kerbl20233d} for scene representation and differentiable image formation. This enables gradient-based optimisation of optics, sensors, and placement while retaining the capacity to model complex, unconventional camera systems beyond the reach of prior differentiable methods. Since 3DGS scenes can be reconstructed from real-world captures, our method trains on both rendered and captured data; discrete parameters such as the coded aperture mask and sensor selection are optimised via a continuous-relaxation surrogate gradient. We demonstrate scalability to high-dimensional design spaces by jointly optimising a coded aperture mask with the remaining camera parameters and a depth-from-defocus perception task, for both single-camera and stereo configurations, analysing the resulting design trade-offs, an analysis made possible by our 3DGS-based representation and largely overlooked in prior coded aperture work. In summary, our contributions are:
\begin{itemize}[noitemsep,topsep=0pt]
\item We introduce a differentiable framework that jointly optimises camera systems and their perception networks using 3DGS scenes reconstructed from real-world captures.
\item We use this framework to design both monocular and stereo coded aperture cameras, via a depth-from-defocus network augmented with a frequency injection module and extended to the stereo setup.
\item We systematically compare against existing camera design approaches and demonstrate significant improvements (e.g., up to 65\% lower all-in-focus (AiF) LPIPS and 13\% lower depth RMSE compared to the strongest baseline in the monocular setup), and analyse the design trade-offs between camera parameters, aperture masks, and task performance.
\end{itemize}

This work provides a unified, differentiable framework that bridges scene representation, image formation, and perception, enabling systematic study of how camera design choices affect downstream task performance.

%% file: sec/2_lit.tex
\section{Related Work}
\label{diff-sec:lit}
\smallskip \noindent \textbf{Task-Driven Camera Design}\hspace{2pt} Designing cameras for perception tasks improves performance, though many approaches rely on non-differentiable simulators. Early work used renderers such as ISET~\cite{Wandell2024ISET3D, Wandell2024ISETCam} to prototype cameras and manually tune parameters against task outcomes~\cite{blasinski2018optimizing, liu2019system, liu2019soft, weikl2021end, liu2023using}, trading laborious manual iteration for lower development cost. Later work automates this with reinforcement-learning~\cite{klinghoffer2023diser, hou2023optimizing} or evolutionary~\cite{yan2025tacos, yan2026joca, tiwary2025if} methods that bypass large networks for faster, more data-efficient optimisation. Treating the simulator as a black box, however, makes these methods inefficient with high-dimensional or numerous parameters such as a coded aperture mask, which our method handles efficiently via gradient-based optimisation while retaining full camera system support.

Another line of work instead leverages differentiable simulation to jointly optimise camera parameters and the perception task, applied to depth of field~\cite{sitzmann2018end, sun2021end, yang2024curriculum}, object detection~\cite{tseng2019hyperparameter, tseng2021differentiable, robidoux2021end, cote2023differentiable}, image classification~\cite{chang2018hybrid, diamond2021dirty, zhang2022all, yang2023image}, depth estimation~\cite{he2018learning, chang2019deep, ikoma2021depth, baek2021polka}, HDR imaging~\cite{metzler2020deep, martel2020neural}, and motion deblurring~\cite{nguyen2022learning}. These methods simulate cameras' effect on pre-captured 2D images, however, limiting generalisation to variable FoV, resolution, unconventional configurations, or multi-camera systems. Our approach instead constructs the full environment in 3D using 3DGS, supporting such scenarios while retaining differentiability.

\smallskip \noindent \textbf{Coded Aperture Imaging}\hspace{2pt} Coded aperture imaging modifies the aperture to encode scene-dependent blur, improving depth and AiF recovery. Early work showed that appropriately designed aperture patterns improve the conditioning of depth-from-defocus and deblurring~\cite{levin2007image, veeraraghavan2007dappled, zhou2009coded}, and recent methods jointly optimise the aperture code and reconstruction network end-to-end using amplitude~\cite{shedligeri2017data} or phase masks~\cite{haim2018depth, wu2019phasecam3d, ikoma2021depth}. Ghanekar et al.~\cite{ghanekar2024passive} extended this to dual-pixel sensors, learning an amplitude mask that trades depth estimation against deblurring.

These methods optimise the aperture mask alone, whereas we treat coded aperture imaging as a broader co-design problem, jointly optimising the mask with other camera parameters for task-driven objectives. This enables systematic exploration of trade-offs such as depth accuracy, noise, and defocus blur, and we further extend coded aperture imaging to stereo-camera systems.

%% file: sec/3_method.tex
\section{Methodology}
\label{diff-sec:method}
\begin{figure*}[t]
        \centering
        \includegraphics[width=\linewidth]{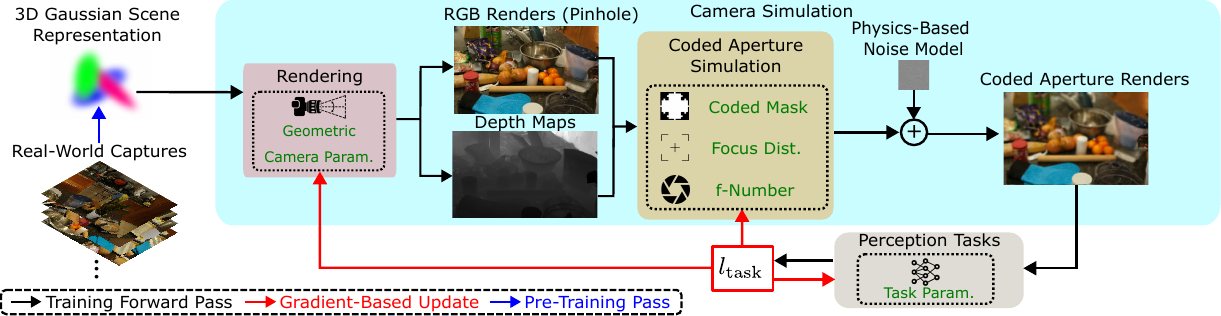}
        \caption{A camera captures the real-world scene (blue arrow), which 3DGS models as a 3D Gaussian representation. At each step, this representation is rendered with the current geometric camera parameters into pinhole RGB images and depth maps; the coded aperture simulation then convolves these RGB renders with depth-dependent PSFs computed from the coded mask, focus distance, and f-number via a physical optics model, followed by a physics-based noise model. The resulting renders are fed to the perception task network, and the task loss $l_{\text{task}}$ is backpropagated through both stages (red arrows) to jointly update all learnable parameters (green text).}
\label{diff-fig:method_overview}
\end{figure*}

We propose a differentiable framework that jointly optimises camera design and downstream perception tasks using a 3DGS scene representation, as illustrated in Figure~\ref{diff-fig:method_overview}.

\subsection{Scene Representation and Image Formation}
We represent the 3D scene using 3DGS~\cite{kerbl20233d}, a differentiable splatting-based representation reconstructed before optimisation from a set of input images, which renders RGB images and depth maps via alpha-compositing. We assume a pinhole projection model, though other camera models could be used; this yields analytical gradients with respect to camera parameters, enabling gradient-based joint optimisation of intrinsics (e.g., focal length, sensor dimensions) and extrinsics (e.g., placement, orientation).

To simulate coded aperture defocus, we apply an occlusion-aware layered blur model adapted from Ikoma et al.~\cite{ikoma2021depth}: the rendered scene is quantised into discrete depth layers, each convolved with a depth-dependent point spread function (PSF) determined by the coded mask, focus distance, and f-number, then composited back-to-front using the depth maps to account for occlusion between layers.

Following~\cite{yan2025tacos}, we augment the rendered image with an affine noise model~\cite{foi2009clipped} combining signal-dependent Poisson noise and thermal noise calibrated on a physical camera. The calibrated noise parameters generalise to arbitrary exposure, gain, pixel size, and aperture size by scaling the effective gain with the ratios of pixel and aperture areas relative to the calibrated configuration, including the coded mask's open-aperture fraction.

\subsection{Perception Task}
Our framework is compatible with any differentiable perception task, but here we focus on an RGB-D task that jointly estimates depth from defocus blur and reconstructs an AiF image, as it depends directly on the optical parameters we optimise. Our monocular network builds on the depth-from-defocus architecture of Ikoma et al.~\cite{ikoma2021depth}, which deconvolves the coded aperture image into a focal stack via a Tikhonov inverse filter and predicts the AiF image and depth map from the stack and coded image using a U-Net backbone. We augment this backbone with a frequency injection module that extracts local defocus cues from patch-wise Fourier spectra of the coded image and injects them at the U-Net bottleneck, improving depth separation.

To demonstrate our method's ability to design beyond a single camera, we extend this network to a stereo setup: we deconvolve each view's coded aperture image into a focal stack as before, then fuse them into intermediate AiF images via a differentiable Laplacian sharpness measure for photometric consistency across views. Inspired by~\cite{choi2025dual}, a frozen pretrained optical flow network, RAPIDFlow~\cite{morimitsu2024rapidflow}, estimates disparity from these images and warps the right focal stack into the left view's coordinate frame. The left and warped right stacks, together with a disparity-derived depth cue, are then fed into the same task network to predict the left view's AiF image and depth map. Illustration diagrams of both setups are in the supplementary material.

\subsection{Optimisation}\label{diff-sec: optimisation}
We formulate our framework as a joint optimisation problem over the camera parameters and the perception task, as illustrated in Figure~\ref{diff-fig:method_overview}. At each step, the 3DGS scene is rendered into RGB and depth images, passed through the coded aperture and noise simulation, and fed to the perception network; the resulting task loss $l_{\text{task}}$ is backpropagated through the entire pipeline to jointly update all camera and network parameters using the Adam optimiser~\cite{kingma2015adam}.

\begin{table*}[!h]
\centering
\caption{Quantitative results for monocular camera design, comparing our joint camera, mask, and perception optimisation against a fixed-camera baseline (Ikoma et al.~\cite{ikoma2021depth}, tested with their original \emph{Phase} mask and with an \emph{Amp.} (amplitude) mask matched to our settings) and a derivative-free joint optimiser (TaCOS~\cite{yan2025tacos}). Optimised parameters are marked \textcolor{green}{$\bigcdot$}, fixed parameters $\bigcdot$; our method outperforms both baselines under two noise scenarios.
}
\resizebox{\textwidth}{!}{%
\begin{tabular}{cc>{\centering\arraybackslash}m{70pt}cccccccccccc}
\hline
\multirow{3}{*}{Scenario} & \multirow{3}{*}{Method} & \multicolumn{6}{c}{Camera Parameters} & \multicolumn{7}{c}{Performance}  \\
\cline{3-15}
&  & Coded Mask & Focus Dist. & f-num & Focal Length & Sensor Size & Pixel Size & \multicolumn{3}{c}{All-in-Focus} & \multicolumn{4}{c}{Depth} \\
& & $M$ & focus ($m$) & $f$ & focal (mm) & $w\times h$  (mm) & $p$ ($\mu$m) & PSNR (dB) $\uparrow$ & SSIM $\uparrow$ & LPIPS $\downarrow$ & RMSE (m) $\downarrow$ & MAE (m) $\downarrow$ & $\text{log}_{10}$ $\downarrow$ & $\delta<1.25$ $\uparrow$ \\
\hline
\multirow{4}{*}[-25pt]{Noise$\times$100} & Fixed Cam. + Phase + Ikoma et al.~\cite{ikoma2021depth} & \includegraphics[height=25pt]{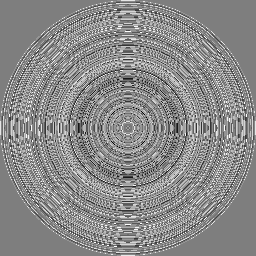} \textcolor{green}{$\bigcdot$} & 1.70 $\bigcdot$ & 1.8 $\bigcdot$ & 14.29 $\bigcdot$ & 14.47$\times$9.94 $\bigcdot$ & 9.00 $\bigcdot$ & 22.587 & 0.669 & 0.308 & 0.131 & \underline{0.080} & 0.102 & 0.910\\
& Fixed Cam. + Amp. + Ikoma et al.~\cite{ikoma2021depth} & \includegraphics[height=25pt]{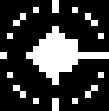} \textcolor{green}{$\bigcdot$} & 1.70 $\bigcdot$ & 1.8 $\bigcdot$ & 14.29 $\bigcdot$ & 14.47$\times$9.94 $\bigcdot$ & 9.00 $\bigcdot$ & 23.543 & 0.696 & \underline{0.292} & \underline{0.126} & \underline{0.080} & \underline{0.098} & \underline{0.917}\\
& TaCOS + Amp. + Proposed Model & \includegraphics[height=25pt]{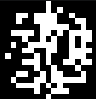} \textcolor{green}{$\bigcdot$} & 0.99 \textcolor{green}{$\bigcdot$} & 11.9 \textcolor{green}{$\bigcdot$} & 24.95 \textcolor{green}{$\bigcdot$} & 14.44$\times$9.90 \textcolor{green}{$\bigcdot$} &  4.50\textcolor{green}{$\bigcdot$} & \underline{24.494} & \underline{0.767} & 0.401 & 0.169 & 0.115 & 0.131 & 0.868\\
& Ours - Optim. Cam. + Amp. + Proposed Model & \includegraphics[height=25pt]{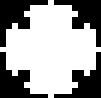} \textcolor{green}{$\bigcdot$} & 0.45 \textcolor{green}{$\bigcdot$} & 12.4 \textcolor{green}{$\bigcdot$} & 12.37 \textcolor{green}{$\bigcdot$} & 14.47$\times$9.94 \textcolor{green}{$\bigcdot$} & 9.00 \textcolor{green}{$\bigcdot$} & \textbf{29.301} & \textbf{0.893} & \textbf{0.102} & \textbf{0.110} & \textbf{0.070} & \textbf{0.091} & \textbf{0.938}\\
\hline
\multirow{4}{*}[-25pt]{Noise$\times$200} & Fixed Cam. + Phase + Ikoma et al.~\cite{ikoma2021depth} & \includegraphics[height=25pt]{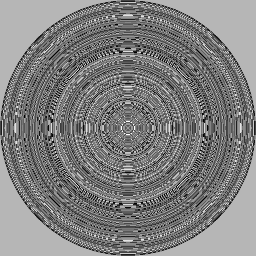} \textcolor{green}{$\bigcdot$} & 1.70 $\bigcdot$ & 1.8 $\bigcdot$ & 14.29 $\bigcdot$ & 14.47$\times$9.94 $\bigcdot$ & 9.00 $\bigcdot$ & 22.325 & 0.668 & 0.318 & 0.133 & \underline{0.078} & \underline{0.102} & \underline{0.912}\\
& Fixed Cam. + Amp. + Ikoma et al.~\cite{ikoma2021depth} & \includegraphics[height=25pt]{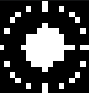} \textcolor{green}{$\bigcdot$} & 1.70 $\bigcdot$ & 1.8 $\bigcdot$ & 14.29 $\bigcdot$ & 14.47$\times$9.94 $\bigcdot$ & 9.00 $\bigcdot$ & 23.531 & 0.700 & \underline{0.289} & \underline{0.128} & 0.082 & 0.103 & \underline{0.912}\\
& TaCOS + Amp. + Proposed Model & \includegraphics[height=25pt]{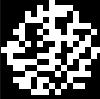} \textcolor{green}{$\bigcdot$} & 1.03 \textcolor{green}{$\bigcdot$} &  9.0 \textcolor{green}{$\bigcdot$} & 24.07 \textcolor{green}{$\bigcdot$} & 14.13$\times$7.45 \textcolor{green}{$\bigcdot$} &  3.45\textcolor{green}{$\bigcdot$} & \underline{23.644} & \underline{0.772} & 0.436 & 0.182 & 0.127 & 0.136 & 0.850\\
& Ours - Optim. Cam. + Amp. + Proposed Model & \includegraphics[height=25pt]{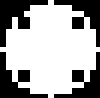} \textcolor{green}{$\bigcdot$} & 0.47 \textcolor{green}{$\bigcdot$} & 11.0 \textcolor{green}{$\bigcdot$} & 14.60 \textcolor{green}{$\bigcdot$} & 14.47$\times$9.94 \textcolor{green}{$\bigcdot$} & 9.00 \textcolor{green}{$\bigcdot$} & \textbf{27.534} & \textbf{0.872} & \textbf{0.116} & \textbf{0.117} & \textbf{0.077} & \textbf{0.099} & \textbf{0.932}\\
\hline
\end{tabular}
}
\label{diff-tab:mono_results}
\end{table*}

\smallskip \noindent \textbf{Camera Parameters}\hspace{2pt} We optimise the optics (focal length, f-number, coded aperture mask, focus distance) and image sensor (dimensions, pixel size) for the monocular setup, plus the baseline between cameras for the stereo setup. Because sparse view coverage near the scene can cause 3DGS renders to contain large black regions under wide FoVs, we apply a black-pixel regularisation loss $\mathcal{L}_{\mathrm{blk}} = \lambda_{\mathrm{blk}} \, \beta / \text{focal}$, where $\beta$ is the rendered black-pixel fraction, $\text{focal}$ is the optimisable focal length, and the weight $\lambda_{\mathrm{blk}}{=}30$ is set empirically to match the scale of the other losses. This loss discourages the optimiser from exploiting short focal lengths where the reconstruction is degenerate; it therefore constrains the effective design space rather than expressing a preference of the downstream perception task, and reflects a limitation of the 3DGS representation's view coverage.

\smallskip \noindent \textbf{Discrete Parameters}\hspace{2pt} The image sensor (chosen from a fixed catalogue for manufacturability) and the coded aperture mask are inherently discrete choices, which we optimise using a Gumbel-Softmax relaxation with a straight-through estimator (STE)~\cite{jang2017categorical}. Given $K$ candidates with logits $\phi_i$, the relaxed selection weights are
\begin{equation}\label{diff-eqn:gumbel-softmax}
    y_i = \frac{\exp\!\left((\phi_i + g_i) / \tau\right)}{\sum_{j=1}^{K} \exp\!\left((\phi_j + g_j) / \tau\right)}, \quad g_i \stackrel{\text{i.i.d.}}{\sim} \text{Gumbel}(0,1),
\end{equation}
where the temperature $\tau$ is annealed towards zero during training so that $\mathbf{y}$ progressively sharpens towards a one-hot vector. The STE uses the hard selection $\arg\max_i(\phi_i + g_i)$ in the forward pass, so the pipeline always renders with a physically realisable discrete configuration, while gradients reach $\boldsymbol{\phi}$ through the soft weights $\mathbf{y}$ in the backward pass.

For sensor selection, $K$ is the catalogue size, and following~\cite{yan2025tacos} we optimise the underlying continuous sensor parameters $\boldsymbol{\theta}$ (pixel size, sensor width, and height) rather than $\boldsymbol{\phi}$ directly, setting $\phi_i$ to the negative normalised squared distance between $\boldsymbol{\theta}$ and catalogue entry $\mathbf{c}_i$. This makes the nearest catalogue entry the most likely selection and implicitly pulls $\boldsymbol{\theta}$ towards the catalogue during optimisation, avoiding the need for an explicit commitment loss. For the coded aperture mask, we apply the same formulation independently to each element with $K=2$ (closed/open).

\smallskip \noindent \textbf{Perception Task}\hspace{2pt} We supervise the predicted AiF image $\hat{I}$ and depth map $\hat{D}$ against ground truth $I$ and $D$ rendered by 3DGS, using a combination of pixel-wise, structural, and colour losses. As $I$ and $D$ are 3DGS renders rather than independent measurements, they may carry the reconstruction's own error; since every method shares the same reference, the resulting metrics remain valid for comparing methods rather than as absolute measures of real-world accuracy. For the monocular network,
\begin{equation}\label{diff-eqn:mono_loss}
\begin{aligned}
    \mathcal{L}_{\mathrm{mono}} \;=\;&
        \lambda_{\mathrm{aif}} \bigl(\|\hat{I}-I\|_1 + \lambda_{\mathrm{g}}\mathcal{L}_{\mathrm{edge}}(\hat{I}, I)\bigr) \\
        &+ \lambda_{\mathrm{d}} \bigl(\|\hat{D}-D\|_1 + \lambda_{\mathrm{g}}\mathcal{L}_{\mathrm{edge}}(\hat{D}, D)\bigr) \\
        &+ \lambda_{\mathrm{ssim}} \bigl(\mathcal{L}_{\mathrm{ssim}}(\hat{I}, I) + \mathcal{L}_{\mathrm{ssim}}(\hat{D}, D)\bigr) \\
        &+ \lambda_{\mathrm{ch}}\,\mathcal{L}_{\mathrm{ch}} + \lambda_{\mathrm{cs}}\,\mathcal{L}_{\mathrm{cs}},
\end{aligned}
\end{equation}
where $\mathcal{L}_{\mathrm{edge}}$ is a Sobel-gradient L1 loss that sharpens edges, $\mathcal{L}_{\mathrm{ssim}}(\hat{y},y)=1-\mathrm{SSIM}(\hat{y},y)$, $\mathcal{L}_{\mathrm{ch}}$ is an L1 loss on the CIE-Lab $a^*b^*$ channels penalising hue/saturation errors, and $\mathcal{L}_{\mathrm{cs}}$ matches per-channel RGB mean and standard deviation to correct colour casts and contrast mismatches. We set $(\lambda_{\mathrm{aif}}, \lambda_{\mathrm{d}}, \lambda_{\mathrm{ssim}}, \lambda_{\mathrm{g}}, \lambda_{\mathrm{ch}}, \lambda_{\mathrm{cs}}) = (1, 2, 1, 1, 0.1, 0.1)$, with the larger depth weight reflecting the greater difficulty of metric depth estimation relative to RGB reconstruction.

For the stereo network, we add a loss between the ground-truth depth and the depth $\bar{D}$ implied by the predicted disparity, giving
\begin{equation}\label{diff-eqn:stereo_loss}
    \mathcal{L}_{\mathrm{stereo}} = \mathcal{L}_{\mathrm{mono}} + \|\bar{D}-D\|_1,
\end{equation}
which also lets gradients optimise the baseline between cameras for disparity accuracy even though the flow network itself is frozen.

\smallskip \noindent \textbf{Total Loss}\hspace{2pt} The total loss for the framework is $\mathcal{L} = \mathcal{L}_{\mathrm{task}} + \mathcal{L}_{\mathrm{blk}}$, where $\mathcal{L}_{\mathrm{task}}$ is the monocular or the stereo loss.

%% file: sec/4_exp.tex
\begin{figure*}[!ht]
        \centering
        \includegraphics[width=\linewidth]{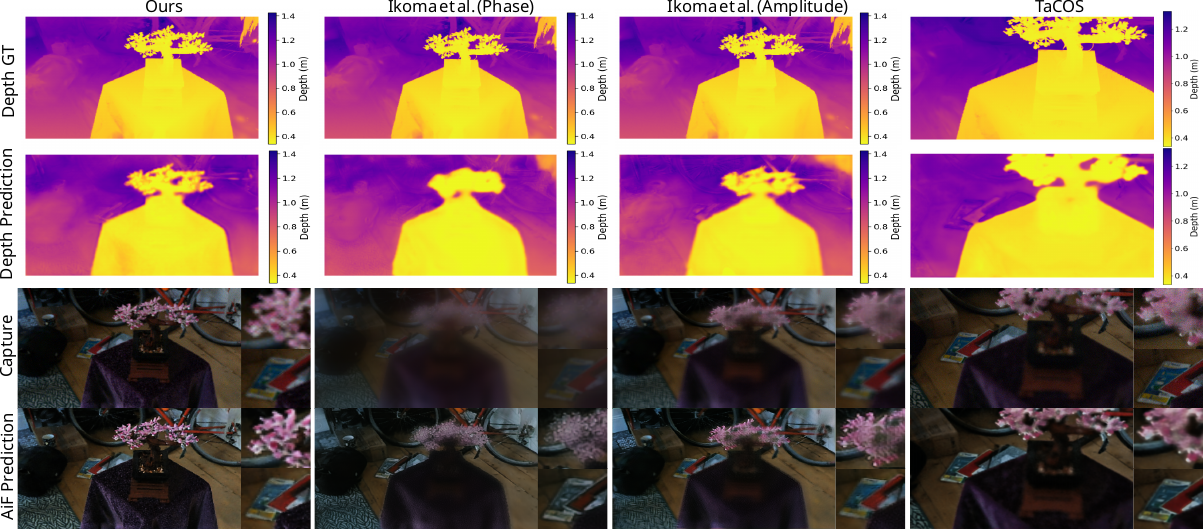}
        \caption{Qualitative comparison for the monocular experiment. Our jointly optimised camera, mask, and perception model produce sharp AiF images and accurate depth, while the fixed-camera baseline (Ikoma et al.~\cite{ikoma2021depth}) and the derivative-free joint optimiser (TaCOS~\cite{yan2025tacos}) degrade noticeably in blurry regions.}
\label{diff-fig:mono_results}
\end{figure*}
\section{Experiments}
\label{diff-sec:exp}
We validate our method on two close-range design scenarios for depth-from-defocus and AiF reconstruction, focusing on short range because the defocus cue is less informative further out. The monocular case targets robotic grasping, recovering depth from a single camera over a confined workspace. The stereo case targets macro photography, where a zoomed-in view with a shallow depth of field (DoF) produces strong defocus blur; here a stereo pair adds complementary photometric and geometric depth cues.

\subsection{Implementations}\label{diff-sec:exp_setup}
\smallskip \noindent \textbf{Dataset and Environment}\hspace{2pt}
We select close-range scenes from MipNeRF~360~\cite{barron2022mip} and LLFF~\cite{mildenhall2019local}: 6 scenes (\emph{leaves}, \emph{flower}, \emph{fortress}, \emph{orchids}, \emph{bonsai}, \emph{room}) for the monocular case, and 7 scenes (the same 4 LLFF scenes plus \emph{counter}, \emph{kitchen}, \emph{bonsai} from MipNeRF~360) for the stereo case. For each scene, we establish a 3D Gaussian representation with gsplat~\cite{ye2025gsplat} as the virtual environment, then recover metric scale by aligning the reconstruction to metric depth from Depth Anything 3~\cite{lin2025depth}. The camera design and task network are jointly optimised across all scenes, with a subset of views per scene withheld for evaluation.

\smallskip \noindent \textbf{Noise Calibration}\hspace{2pt}
We calibrate the noise model on a Basler Dart DaA1280-54uc camera~\cite{Basler2024} (5\,ms exposure, 10\,dB gain, f/8) following~\cite{yan2025tacos}. This configuration is treated as the reference point for generalisation.

\smallskip \noindent \textbf{Optimisation}\hspace{2pt}
All camera parameters, coded aperture masks, and task networks are jointly optimised with Adam~\cite{kingma2015adam} for 20{,}000 steps with a batch size of 3, using a learning rate of $10^{-2}$ for the camera parameters and $10^{-3}$ for the mask and task networks.

\subsection{Monocular Camera}
The monocular camera design task focuses on close-range RGB-D imaging from a single capture, which is a common requirement in robotics applications such as grasping.
\begin{table*}[!ht]
\centering
\caption{Quantitative results for stereo camera design. The monocular row is our own strongest monocular configuration evaluated under the same narrow FoV, isolating the benefit of a second view; CADS~\cite{ghanekar2024passive} is the closest existing coded-aperture baseline, included despite its dual-pixel baseline being limited to the sub-mm aperture scale. Optimised parameters are marked \textcolor{green}{$\bigcdot$}, fixed $\bigcdot$; stereo values are given as \emph{Left}/\emph{Right} (CADS shares parameters across views). Our stereo method outperforms both, especially in AiF reconstruction.
}
\resizebox{\textwidth}{!}{%
\begin{tabular}{cc>{\centering\arraybackslash}m{120pt}cccccccccc}
\hline
\multirow{3}{*}{Scenario} & \multirow{3}{*}{Method} & \multicolumn{4}{c}{Camera Parameters} & \multicolumn{7}{c}{Performance}  \\
\cline{3-13}
&  & Coded Mask & Focus Dist. & f-num & Baseline & \multicolumn{3}{c}{All-in-Focus} & \multicolumn{4}{c}{Depth} \\
& & $M$ & focus ($m$) & $f$ & $b$ (m) & PSNR (dB) $\uparrow$ & SSIM $\uparrow$ & LPIPS $\downarrow$ & RMSE (m) $\downarrow$ & MAE (m) $\downarrow$ & $\text{log}_{10}$ $\downarrow$ & $\delta<1.25$ $\uparrow$ \\
\hline
\multirow{3}{*}[-15pt]{Noise$\times$100} & Fixed Cam. + CADS~\cite{ghanekar2024passive} & \includegraphics[height=25pt]{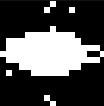} \textcolor{green}{$\bigcdot$} & 0.29 $\bigcdot$ & 9.87 $\bigcdot$ & NA  & \underline{26.397} & 0.759 & 0.301 & 0.244 & 0.135 & 0.188 & 0.797\\
& Ours - Monocular & \includegraphics[height=25pt]{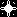} \textcolor{green}{$\bigcdot$} & 0.29 \textcolor{green}{$\bigcdot$} & 16.0 \textcolor{green}{$\bigcdot$} & NA & 26.335 & \underline{0.819} & \underline{0.205} & \underline{0.089} & \underline{0.058} & \underline{0.114} & \underline{0.912}\\
& Ours - Stereo & \includegraphics[height=25pt]{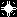} \includegraphics[height=25pt]{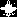} \textcolor{green}{$\bigcdot$} & 0.26/0.89 \textcolor{green}{$\bigcdot$} & 16.0/11.3 \textcolor{green}{$\bigcdot$} & 0.01 \textcolor{green}{$\bigcdot$} & \textbf{26.828} & \textbf{0.841} & \textbf{0.161} & \textbf{0.084} & \textbf{0.049} & \textbf{0.086} & \textbf{0.933} \\
\hline
\multirow{3}{*}[-15pt]{Noise$\times$200} & Fixed Cam. + CADS~\cite{ghanekar2024passive} & \includegraphics[height=25pt]{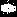} \textcolor{green}{$\bigcdot$} & 0.28 $\bigcdot$ &  9.87 $\bigcdot$ & NA & 25.618 & 0.718 & 0.350 & 0.259 & 0.143 & 0.203 & 0.775\\
& Ours - Monocular & \includegraphics[height=25pt]{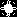} \textcolor{green}{$\bigcdot$} & 0.28 \textcolor{green}{$\bigcdot$} & 16.0\textcolor{green}{$\bigcdot$} & NA & \underline{25.678} & \underline{0.788} & \underline{0.239} & \underline{0.089} & \textbf{0.051} & \underline{0.115} & \underline{0.913}\\
& Ours - Stereo & \includegraphics[height=25pt]{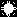} \includegraphics[height=25pt]{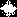} \textcolor{green}{$\bigcdot$} & 0.26/0.86\textcolor{green}{$\bigcdot$} & 16.0/11.1\textcolor{green}{$\bigcdot$} & 0.01\textcolor{green}{$\bigcdot$} & \textbf{26.268} & \textbf{0.813} & \textbf{0.203} & \textbf{0.081} & \textbf{0.051} & \textbf{0.093} & \textbf{0.928} \\
\hline
\end{tabular}
}
\label{diff-tab:stereo_results}
\end{table*}
\begin{figure*}
        \centering
        \includegraphics[width=\linewidth]{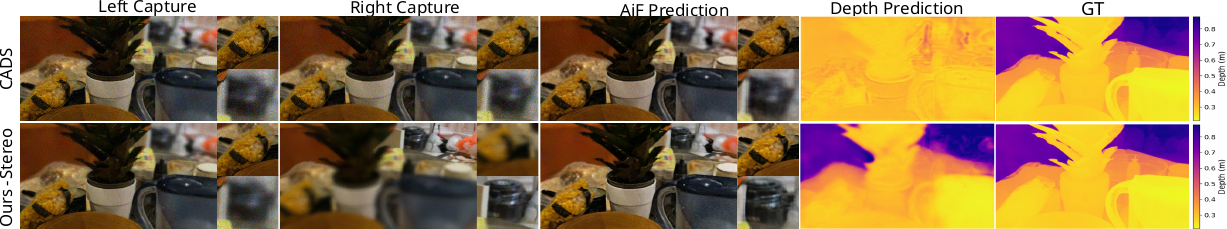}
        \caption{Our method fuses photometric cues from a foreground-focused and a background-focused view for higher AiF quality and improves depth accuracy by leveraging disparity. Full comparisons with baselines are presented in the supplementary material.
        }
\label{diff-fig:stereo_results}
\end{figure*}

\smallskip \noindent \textbf{Design Space}\hspace{2pt} The design space covers the optical parameters (focal length $\in[5,25]$\,mm, focus distance $\in[0.1,5]$\,m, f-number $\in[1.4,16]$, and a $17\times17$\,px circular amplitude coded aperture mask) and the image sensor, chosen from a catalogue of 43 commercial CMOS sensors collected in~\cite{yan2025tacos}. Sensor size and focal length determine the FoV, while focal length and f-number determine the aperture size. The mask and sensor choice are optimised with the Gumbel-Softmax and STE formulation, as in Section~\ref{diff-sec: optimisation}.

\smallskip \noindent \textbf{Comparison}\hspace{2pt} We compare against TaCOS~\cite{yan2025tacos}, a genetic algorithm-based camera design method that jointly optimises camera parameters and the perception model, run for the same $\sim$20{,}000-step budget (590 candidates over 34 generations, following its recommended $2\times$-parameter population size for our 295 optimised parameters).

We also compare to the baseline depth-from-defocus method of Ikoma et al.~\cite{ikoma2021depth} with a fixed camera, testing both their original phase mask, matched in element count, and an amplitude mask matched to our own settings. We use the catalogue's largest-pixel sensor for the best SNR, and reuse their focus distance as it targets a similar depth range to our scenes. We scale the f-number and focal length to reproduce their aperture diameter, matching their level of defocus blur while keeping a reasonable FoV with our sensor. These camera parameters are listed in Table~\ref{diff-tab:mono_results}.

To highlight the benefit of a coded aperture, which recovers an AiF image from the blur caused by widening the aperture under low-light conditions, we evaluate all methods under two low-SNR scenarios with increased image noise, representing low-light operation or noisier sensors.

AiF quality is measured with PSNR, SSIM, and the perceptual metric LPIPS~\cite{zhang2018unreasonable}; depth accuracy is measured with RMSE, MAE, $\text{log}_{10}$ error, and the threshold accuracy $\delta < 1.25$ that measures the fraction of predicted depths within a factor of 1.25 of the ground truth.
\begin{table*}
\centering
\caption{Ablation study under Noise$\times$100. Monocular: a fixed human-designed mask~\cite{levin2007image}, a fixed camera (Ikoma et al.~\cite{ikoma2021depth} configuration), and no frequency injection. Stereo: shared camera parameters, a shared mask, a conventional (uncoded) circular aperture, and no right-view warping. Optimised parameters are marked \textcolor{green}{$\bigcdot$}, fixed $\bigcdot$; `--' marks a parameter not applicable to that scenario. The results confirm that each component or design choice is a meaningful contributor to the proposed design.
}
\resizebox{\textwidth}{!}{%
\begin{tabular}{c>{\centering\arraybackslash}m{55pt}cccccccccccccc}
\hline
\multirow{3}{*}{Scenario} & \multirow{3}{*}{Method} & \multicolumn{7}{c}{Camera Parameters} & \multicolumn{7}{c}{Performance}  \\
\cline{3-16}
& & Coded Mask & Focus Dist. & f-num & Focal Length & Sensor Size & Pixel Size & Baseline & \multicolumn{3}{c}{All-in-Focus} & \multicolumn{4}{c}{Depth} \\
& & $M$ & focus ($m$) & $f$ & focal (mm) & $w\times h$ (mm) & $p$ ($\mu$m) & $b$ (m) & PSNR (dB) $\uparrow$ & SSIM $\uparrow$ & LPIPS $\downarrow$ & RMSE (m) $\downarrow$ & MAE (m) $\downarrow$ & $\text{log}_{10}$ $\downarrow$ & $\delta<1.25$ $\uparrow$ \\
\hline
\multirow{4}{*}[-25pt]{Monocular} & Fixed Mask & \includegraphics[height=25pt]{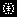} $\bigcdot$ & 0.56\textcolor{green}{$\bigcdot$} & 5.0 \textcolor{green}{$\bigcdot$} & 12.96 \textcolor{green}{$\bigcdot$} & 14.47$\times$9.94 \textcolor{green}{$\bigcdot$} & 9.00 \textcolor{green}{$\bigcdot$} & -- & 27.575 & 0.838 & 0.153 & \underline{0.124} & \underline{0.079} & 0.098 & \underline{0.925}\\
& Fixed Camera & \includegraphics[height=25pt]{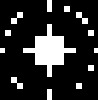} \textcolor{green}{$\bigcdot$} & 1.70 $\bigcdot$ & 1.8 $\bigcdot$ & 14.29 $\bigcdot$ & 14.47$\times$9.94 $\bigcdot$ & 9.00 $\bigcdot$ & -- & 24.397 & 0.740 & 0.290 & \underline{0.124} & 0.081 & \underline{0.095} & 0.920 \\
& No Freq. & \includegraphics[height=25pt]{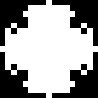} \textcolor{green}{$\bigcdot$} & 0.45 \textcolor{green}{$\bigcdot$} & 12.4 \textcolor{green}{$\bigcdot$} & 14.87 \textcolor{green}{$\bigcdot$} &14.47$\times$9.94 \textcolor{green}{$\bigcdot$} & 9.00\textcolor{green}{$\bigcdot$} & -- & \underline{29.123} & \textbf{0.895} & \textbf{0.100} & 0.125 & 0.081 & 0.101 & \underline{0.925}\\
& Proposed & \includegraphics[height=25pt]{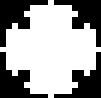} \textcolor{green}{$\bigcdot$} & 0.45 \textcolor{green}{$\bigcdot$} & 12.4 \textcolor{green}{$\bigcdot$} & 12.37 \textcolor{green}{$\bigcdot$} & 14.47$\times$9.94 \textcolor{green}{$\bigcdot$} & 9.00 \textcolor{green}{$\bigcdot$} & -- & \textbf{29.301} & \underline{0.893} & \underline{0.102} & \textbf{0.110} & \textbf{0.070} & \textbf{0.091} & \textbf{0.938}\\
\hline
\multirow{4}{*}[-45pt]{Stereo} & Shared Camera & \includegraphics[height=25pt]{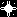} \includegraphics[height=25pt]{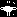} \textcolor{green}{$\bigcdot$} & 0.29/0.29 \textcolor{green}{$\bigcdot$} & 16.0/16.0 \textcolor{green}{$\bigcdot$} & 25.00 $\bigcdot$ & 14.47$\times$9.94 $\bigcdot$ & 9.00 $\bigcdot$ & 0.01 \textcolor{green}{$\bigcdot$}  & 26.123 & 0.815 & 0.223 & 0.101 & 0.055 & 0.099 & 0.913\\
& Shared Mask & \includegraphics[height=25pt]{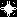} \includegraphics[height=25pt]{masks/stereo_ab/share_mask.png}\textcolor{green}{$\bigcdot$} & 0.26/0.83\textcolor{green}{$\bigcdot$} & 16.0/10.0\textcolor{green}{$\bigcdot$} & 25.00 $\bigcdot$ & 14.47$\times$9.94 $\bigcdot$ & 9.00 $\bigcdot$ & 0.01\textcolor{green}{$\bigcdot$} & \underline{26.662} & \underline{0.832} & \underline{0.177} & \textbf{0.083} & \textbf{0.049} & \underline{0.089} & \underline{0.931}\\
& Circular Aperture & \includegraphics[height=25pt]{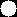} \includegraphics[height=25pt]{masks/stereo_ab/circular.png}$\bigcdot$ & 0.27/0.94\textcolor{green}{$\bigcdot$} & 16.0/11.5\textcolor{green}{$\bigcdot$} & 25.00 $\bigcdot$ & 14.47$\times$9.94 $\bigcdot$ & 9.00 $\bigcdot$ & 0.01\textcolor{green}{$\bigcdot$} & 25.423 & 0.781 & 0.236 & 0.091 & 0.056 & 0.091 & 0.921\\
& w/o Warping & \includegraphics[height=25pt]{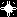} \includegraphics[height=25pt]{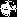} \textcolor{green}{$\bigcdot$} & 0.27/1.48\textcolor{green}{$\bigcdot$} & 16.0/7.13\textcolor{green}{$\bigcdot$} & 25.00 $\bigcdot$ & 14.47$\times$9.94 $\bigcdot$ & 9.00 $\bigcdot$ & 0.01\textcolor{green}{$\bigcdot$} & 26.567 & 0.826 & 0.205 & \underline{0.084} & 0.051 & 0.093 & 0.926\\
& Proposed & \includegraphics[height=25pt]{masks/stereo_100/stereo_l.png} \includegraphics[height=25pt]{masks/stereo_100/stereo_r.png} \textcolor{green}{$\bigcdot$} & 0.26/0.89 \textcolor{green}{$\bigcdot$} & 16.0/11.3 \textcolor{green}{$\bigcdot$} & 25.00 $\bigcdot$ & 14.47$\times$9.94 $\bigcdot$ & 9.00 $\bigcdot$ & 0.01 \textcolor{green}{$\bigcdot$} & \textbf{26.828} & \textbf{0.841} & \textbf{0.161} & \underline{0.084} & \textbf{0.049} & \textbf{0.086} & \textbf{0.933} \\
\hline
\end{tabular}
}
\label{diff-tab:ablation}
\end{table*}

\smallskip \noindent \textbf{Results}\hspace{2pt}
Table~\ref{diff-tab:mono_results} reports the camera parameters, coded aperture masks, and metrics for each method under both scenarios, with qualitative results for the Noise$\times$100 scenario in Figure~\ref{diff-fig:mono_results} and Noise$\times$200 in the supplementary material.

Ikoma et al.~\cite{ikoma2021depth}'s fixed camera has a large aperture diameter, producing strong defocus blur across the scene depth range that severely limits AiF quality, as the task model must invert a poorly conditioned PSF with little residual texture. The amplitude mask compensates by concentrating transmittance near the aperture centre to shrink the PSF spread, while keeping sparse coding at the boundary to retain depth cues. AiF quality suffers, but depth metrics stay competitive since the large aperture still yields strong defocus cues. Under higher noise, both variants degrade only marginally, as the mask opens more pixels to maintain light throughput.

TaCOS~\cite{yan2025tacos} selects a long focal length and high f-number. This reduces defocus blur relative to Ikoma et al.~\cite{ikoma2021depth} and modestly improves AiF PSNR/SSIM, but LPIPS and depth accuracy remain substantially worse than others. We attribute this to two factors: TaCOS represents the mask as a 1D array, preventing it from learning 2D spatial coding patterns and hurting LPIPS in particular, and its derivative-free optimiser lacks gradient information to resolve interactions among the many camera parameters. Under Noise$\times$200, it lowers the f-number to widen the aperture, but also selects a smaller pixel size sensor that offsets the SNR gain, indicating that TaCOS struggles to find globally consistent configurations at this parameter scale, though more generations may help. It also takes substantially longer to optimise (3.6\,h vs.\ our 2.1\,h on a NVIDIA RTX PRO 6000 GPU) due to the smaller pixel sizes it selects and its optimisation method.

Our method converges to a short focal length and high f-number, giving a small aperture, $\approx$1.0\,mm under Noise$\times$100, and correspondingly large DoF that keeps the PSF well-conditioned across the workspace, enabling accurate AiF reconstruction while retaining enough residual defocus for the best depth results across all metrics. It also selects the largest available pixel size to maximise SNR at this small aperture, and the mask compensates for the small aperture area by opening more pixels to maintain light throughput. Under Noise$\times$200, it modestly increases the focal length and lowers the f-number, enlarging the aperture to $\approx$1.3\,mm to admit more light while keeping the same sensor. Despite the resulting increase in blur, it sustains strong performance and outperforms all baselines by a substantial margin in both scenarios. Note that the black-pixel regularisation loss discourages the optimiser from selecting even shorter focal lengths, so the converged values reported here reflect this constrained design space rather than an unconstrained preference of the perception task.

\subsection{Stereo Camera}
The stereo experiment targets a close-range, small-FoV macro-photography setting, aiming for a camera pair whose complementary photometric and geometric cues improve on the monocular baseline.

\smallskip \noindent \textbf{Design Space}\hspace{2pt} We fix the focal length (25\,mm) and sensor ($w{=}14.47$, $h{=}9.94$\,mm, $p{=}9.00\,\mu$m, from the monocular catalogue) identically for both cameras, giving a zoomed-in, high-SNR view and the matched FoV needed for reliable disparity estimation. Each camera independently optimises its focus distance ($\in[0.1,5]$\,m), f-number ($\in[1.4,16]$), and mask as in the monocular setup, plus a shared stereo baseline ($b\in[0.01,1.00]$\,m).

\smallskip \noindent \textbf{Comparison}\hspace{2pt} We compare against our own monocular design (under the same FoV), the strongest method from the monocular experiment, and against CADS~\cite{ghanekar2024passive}, a dual-pixel method that jointly learns a coded mask with a depth-estimation task. A dual-pixel camera is conceptually similar to a stereo pair in that it captures two views from slightly different perspectives by splitting the aperture into two halves, each projecting onto a separate photodiode within each pixel, but its effective baseline is limited to the aperture diameter rather than a physical camera separation. We match its focal length, sensor, and aperture diameter (via f-number) to our setup, but use our optimised focus distance in place of its own, as the original setting targets an extremely shallow depth range unsuitable for our scenes. All experiments use the same low-SNR scenarios and metrics as the monocular study.

\begin{figure*}[ht!]
        \centering
        \includegraphics[width=\linewidth]{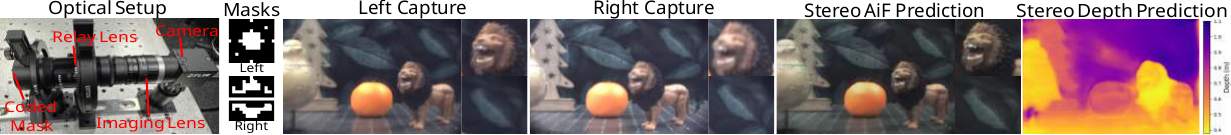}
        \caption{Prototype of the optical system used for real-world validation and the real-world stereo results.
        }
\label{diff-fig:real_world_main}
\end{figure*}

\smallskip \noindent \textbf{Results}\hspace{2pt}
Table~\ref{diff-tab:stereo_results} and Figure~\ref{diff-fig:stereo_results} report the results; full qualitative comparisons are in the supplementary material.

Our method converges to a complementary two-view design: the left camera takes the maximum f-number ($f{=}16.0$, aperture $\approx$1.56\,mm) and a short focus distance (0.26\,m), maximising DoF for the sharpest possible AiF base images, while the right camera focuses further out (0.89\,m) at a lower f-number ($\approx$11.3, aperture $\approx$2.2\,mm), adding depth cues from a different focal plane. The optimised baseline is very small ($b{=}10$\,mm), which is enough for accurate right-to-left warping and photometric fusion, but not larger, since defocus already provides strong geometric depth cues at this range. This complementary design consistently improves on the monocular baseline, most notably in perceptual quality and depth accuracy. The masks follow the same logic: the left mask mirrors the monocular one (small central opening, sparse coded periphery, since it carries the primary supervision signal), while the right mask opens more, and asymmetrically towards the side away from the left camera. This is the region occluded from the left view, where the optimiser raises aperture weight to capture complementary information. Under Noise$\times$200, both masks open further to compensate for SNR, and the stereo method retains its AiF and depth advantage over monocular. The MAE gain vanishes, however, as the extra noise degrades the deconvolved focal stack feeding the disparity estimator, weakening the right-to-left warp and depth prior.

CADS~\cite{ghanekar2024passive} matches our monocular AiF PSNR but is markedly worse in perceptual quality, indicating it preserves mean intensity at the expense of texture. This also shows that our Tikhonov-deconvolution step aids AiF reconstruction. Its depth accuracy is considerably weaker, which we attribute to a depth-range mismatch: CADS is trained on defocus size for depth, and targets millimetre-scale scenes (e.g.\ endoscopy) where defocus size varies monotonically with depth. Our 0.1--5\,m range instead makes this relationship non-linear and ambiguous near the focus plane, so supervising on defocus size no longer yields accurate depth, and its dual-pixel baseline is also too small to compensate geometrically. Its mask reflects this: a large open centre with sparse peripheral coding that splits the aperture to encode disparity rather than modulating spatial frequency, limiting its coding richness relative to our masks that are free across the full aperture area.

\subsection{Ablation Study}
We isolate the contribution of key design choices in both experiments under Noise$\times$100. Results are in Table~\ref{diff-tab:ablation}.

For the monocular method, fixing the mask to the human-designed pattern of Levin et al.~\cite{levin2007image} makes the camera optimiser widen the aperture to compensate for its lower open fraction. Performance still slightly degrades across all metrics, however, demonstrating that optimising camera parameters alone cannot recover the spatial structure of a jointly optimised 2D mask. Fixing the camera to the Ikoma et al.~\cite{ikoma2021depth} configuration substantially reduces AiF quality, confirming that its large aperture and shallow DoF limit reconstruction. It still outperforms the equivalent fixed-camera baseline from the main comparison, however, showing that our model and losses help even without camera optimisation. Removing the frequency injection module (reverting to the Ikoma et al.\ backbone) leaves AiF quality largely unchanged but degrades depth accuracy. The optimiser compensates with a longer focal length to increase blur, confirming the module mainly aids depth estimation via depth-dependent defocus features.

For the stereo method, sharing camera parameters across views collapses both to the left-camera design, removing the complementary focus split and degrading both AiF and depth. The right-view mask compensates by becoming horizontally anisotropic, encoding directional defocus cues since it can no longer differentiate through focus or aperture. Sharing a mask across views causes a modest AiF drop (LPIPS 0.177 vs.\ 0.161) but slightly improves depth (RMSE 0.083 vs.\ 0.084), as identical defocus patterns give more photometrically consistent input to the disparity estimator. Replacing the coded mask with a conventional open aperture, reducing to a standard stereo camera, significantly degrades AiF reconstruction (LPIPS 0.236 vs.\ 0.161): an uncoded circular aperture produces a PSF that is poorly conditioned for deconvolution, so sharpness cannot be recovered through reconstruction the way it can with the coded pattern. Depth accuracy degrades mildly, as stereo disparity still recovers depth correctly over most of the scene, though the weaker AiF reconstruction locally blurs the depth prediction near edges. Removing the warping stage pushes the right camera to a much longer focus distance and wider aperture, creating stronger and more spatially distinctive blur that can be exploited without requiring pixel-level correspondence. This only mildly hurts depth accuracy but degrades AiF (LPIPS 0.205 vs.\ 0.161). The mask also opens more on the side overlapping the left camera's FoV, the region of highest cross-view photometric consistency it can still exploit without explicit correspondence.

\subsection{Real-World Validation}
We further validate our method on a prototype built from a commercial FLIR Flea3 camera~\cite{FLIR2017} with a C-mount lens (Figure~\ref{diff-fig:real_world_main}) due to manufacturing constraints. Hence, this validation focuses on optimising the coded mask, focus distance, and perception network, with the remaining optics fixed. The test scene's 3DGS representation is added to the training set, though the evaluated view itself is not part of the training views. We validate both the monocular and stereo settings, with setup details, PSF calibrations, and monocular results reported in the supplementary material. The stereo camera is achieved by moving the camera by the baseline distance, and the corresponding results are shown in Figure~\ref{diff-fig:real_world_main}. The real-world behaviour matches our simulation: the stereo method focuses each view at a different depth to fuse complementary cues for AiF reconstruction, and the recovered depth aligns with the true scene geometry, confirming the method's applicability beyond simulation.

%% file: sec/5_conclusion.tex
\section{Conclusion}
\label{diff-sec:conclusion}
We presented a differentiable framework that jointly optimises camera design and downstream perception tasks, using 3D Gaussian representations reconstructed from real captures as a differentiable virtual environment for image formation. This avoids the scalability limits of derivative-free optimisation. We validated the framework on monocular and stereo depth-from-defocus tasks with jointly optimised coded apertures, sensors, and inter-camera geometry. It consistently outperforms fixed-camera and derivative-free baselines, while revealing interpretable design trade-offs that a decoupled process would miss. We believe this is an important step toward principled, automated camera design for autonomous systems that accounts for the interdependency between cameras and the algorithms that interpret them. For future work, we aim to extend the framework to jointly optimise camera count and array topology, and to incorporate adaptive, task-driven control of camera parameters online at inference time.

%% file: supplementary.tex

\twocolumn[{%
    \centering
    \vspace{1.5em}
    \includegraphics[width=\linewidth]{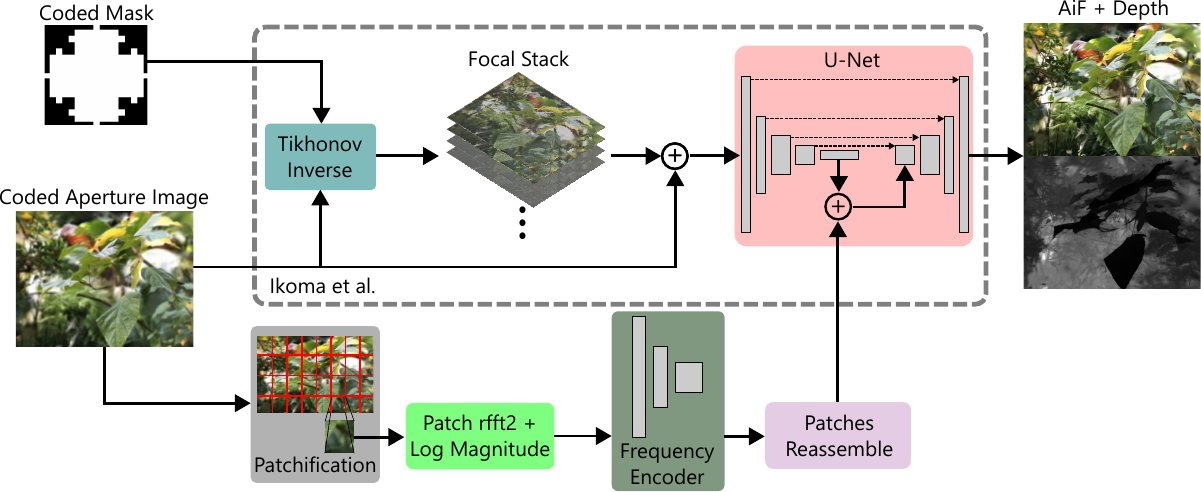}\\
    \vspace{0.5em}
    \captionof{figure}[Monocular depth-from-defocus method]{The monocular depth-from-defocus method. The backbone follows the U-Net of Ikoma et al.~\cite{ikoma2021depth}, which takes a coded aperture image and its Tikhonov-deconvolved focal stack as input and jointly predicts the AiF image and depth map. We augment this backbone with a frequency injection module. In this module, the input image is divided into patches, a 2D Fourier transform is applied to each patch independently, and the log-magnitude spectrum is computed to capture local, spatially-varying defocus strength. The per-patch spectra are encoded by a lightweight 3-layer convolutional network, reassembled into a 2D feature map based on each patch's location, and injected at the bottleneck of the U-Net.}
    \label{diff-fig:mono_model}
    \vspace{1em}
}]

\begin{figure*}[t]
        \centering
        \includegraphics[width=\linewidth]{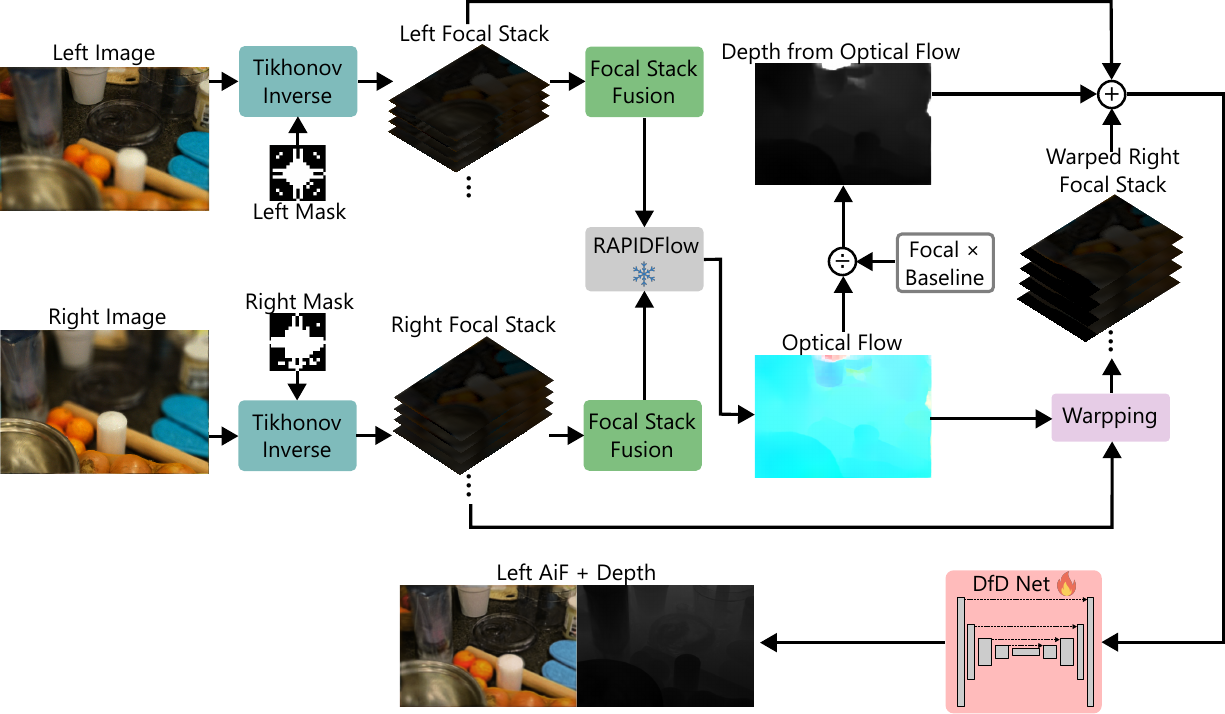}
        \caption[Stereo depth-from-defocus method]{The stereo depth-from-defocus method. Per-view AiF proxy images are synthesised by fusing the Tikhonov-deconvolved focal stack using a differentiable Laplacian sharpness measure and passed to a frozen RAPIDFlow~\cite{morimitsu2024rapidflow} network to estimate dense disparity. The disparity is then used to warp the right-view into the left-view coordinate frame via bilinear grid sampling. The left-view focal stack, the warped right-view focal stack, and a depth map derived from the disparity estimate are concatenated and fed into the monocular U-Net backbone with frequency injection, yielding the reconstructed AiF image and depth map for the left view.}
\label{diff-fig:stereo_model}
\end{figure*}

\section{Depth-from-Defocus Networks}
We provide additional details of the depth-from-defocus networks used as perception tasks in our experiments. The structures of both the monocular and stereo versions are described in this section.

\subsection{Monocular}
As illustrated in Figure~\ref{diff-fig:mono_model}, our monocular task network builds upon the depth-from-defocus architecture of~\cite{ikoma2021depth}. The original method first applies a Tikhonov inverse filter to the rendered coded aperture images, using the PSFs to deconvolve them into a focal stack; each plane of this stack contains a relatively sharp region corresponding to that plane's PSF. However, because the forward model for defocus blur simulation is non-linear, the Tikhonov inverse cannot directly invert this process to obtain the AiF image. The focal stack, together with the rendered coded aperture image, is therefore used as input to a convolutional neural network with a U-Net architecture, which outputs both the AiF image and depth map. Details of this architecture can be found in~\cite{ikoma2021depth}.

We modify this model by incorporating a frequency injection module, as shown in Figure~\ref{diff-fig:mono_model}, that encodes spatially-local defocus cues directly from the coded image in the frequency domain for better depth separation as in~\cite{levin2007image}. Rather than transforming the entire image globally, we follow~\cite{shedligeri2017data} and operate on local patches. The rendered blurry image $\mathbf{b}$ is partitioned into overlapping $P \times P$ windows with stride $S$ via reflection-padded unfolding, yielding $N_h \times N_w$ patches $\{\mathbf{b}_{ij}\}$. Each patch is independently transformed via the real-valued 2D Fourier transform and its log-magnitude spectrum $\log(1 + |\mathcal{F}\{\mathbf{b}_{ij}\}|)$ is computed to obtain local defocus cues. A lightweight 3-layer convolutional encoder with shared weights maps each patch spectrum to a descriptor via global average pooling; these are then reassembled into a dense feature map preserving the spatial correspondence between each descriptor and its originating image region. This map is then injected into the U-Net bottleneck via a learned per-channel gated addition:
\begin{equation}
    \mathbf{x} \leftarrow \mathbf{x} + \sigma(\mathbf{g}) \odot \mathbf{f}_{\mathrm{freq}},
\end{equation}
where $\mathbf{g}$ is initialised to $-3$ so that the branch contributes negligibly at the start of training and the network learns the appropriate mixing ratio. Fusing at the bottleneck allows the spatially-structured frequency cues to condition the entire decoder without perturbing the fine-grained spatial skip connections.

\subsection{Stereo}
For the stereo camera setup, we reuse the U-Net architecture with frequency injection from the monocular version as the base model, with modifications to its input. An illustration is displayed in Figure~\ref{diff-fig:stereo_model}.

We first estimate the disparity between the two views with a pretrained optical flow network, RAPIDFlow~\cite{morimitsu2024rapidflow}, following~\cite{choi2025dual}; the network remains frozen during training. To obtain reliable optical flow estimates, we first construct an intermediate all-in-focus image per view by fusing the Tikhonov-inverted focal stack via a differentiable Laplacian sharpness measure, which improves photometric consistency across the two views. The fused images are then passed to RAPIDFlow to predict a dense disparity map, which warps the right-view focal stack into the left view's coordinates via bilinear grid sampling. This provides additional photometric and defocus cues for AiF reconstruction and depth estimation. The left focal stack, the warped right focal stack, and a depth map derived from the disparity estimate are used as inputs to the same U-Net architecture as in the monocular setting, with the input shape modified accordingly; the network outputs the reconstructed AiF image and depth map for the left view.

\begin{figure*}[!ht]
        \centering
        \includegraphics[width=\linewidth]{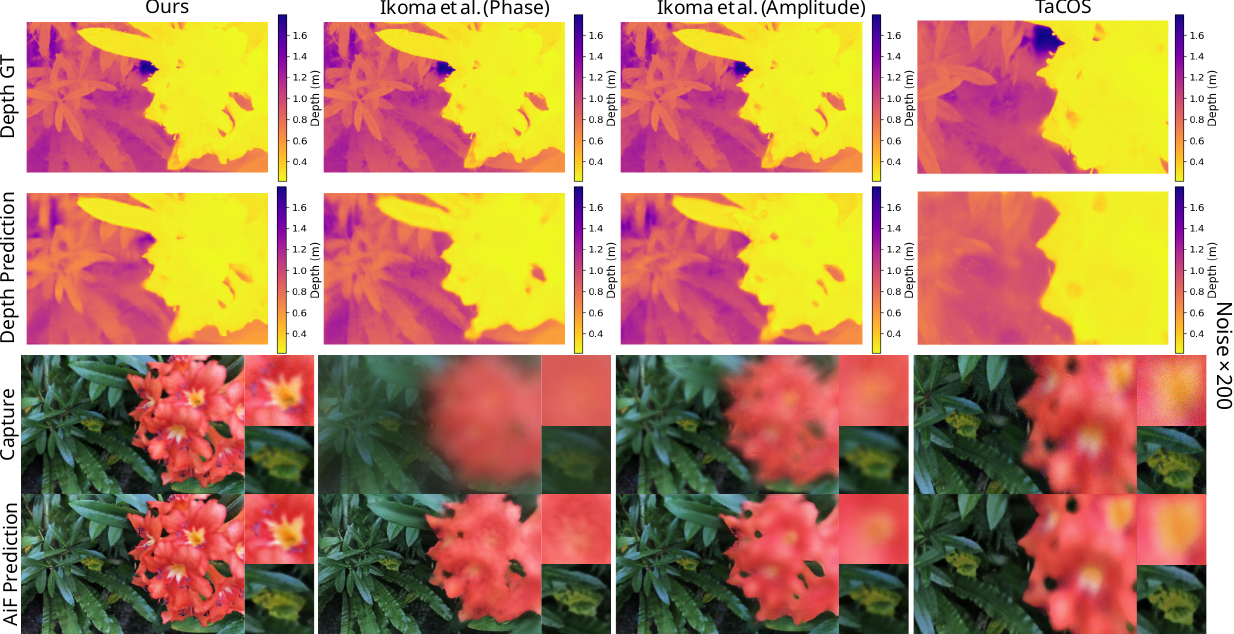}
        \caption{Qualitative comparison for the monocular experiment under the Noise$\times$200 scenario. Our jointly optimised camera, mask, and perception model again produce sharp AiF images and accurate depth, while the fixed-camera baseline (Ikoma et al.~\cite{ikoma2021depth}) and the derivative-free joint optimiser (TaCOS~\cite{yan2025tacos}) produce images with more defocus blur and lower depth accuracy in these regions.}
\label{diff-fig:mono_200}
\end{figure*}

\section{Additional Qualitative Results on Monocular Camera Design}
We provide further qualitative comparisons for the monocular camera design experiment under the lower-SNR (Noise$\times$200) scenario in Fig.~\ref{diff-fig:mono_200}. These results are consistent with the Noise$\times$100 scenario reported in the main paper: our method achieves better AiF reconstruction and depth estimation than the existing baselines. This further supports the effectiveness of jointly optimising camera hardware and perception under challenging noise conditions.

\begin{figure*}[!ht]
        \centering
        \includegraphics[width=\linewidth]{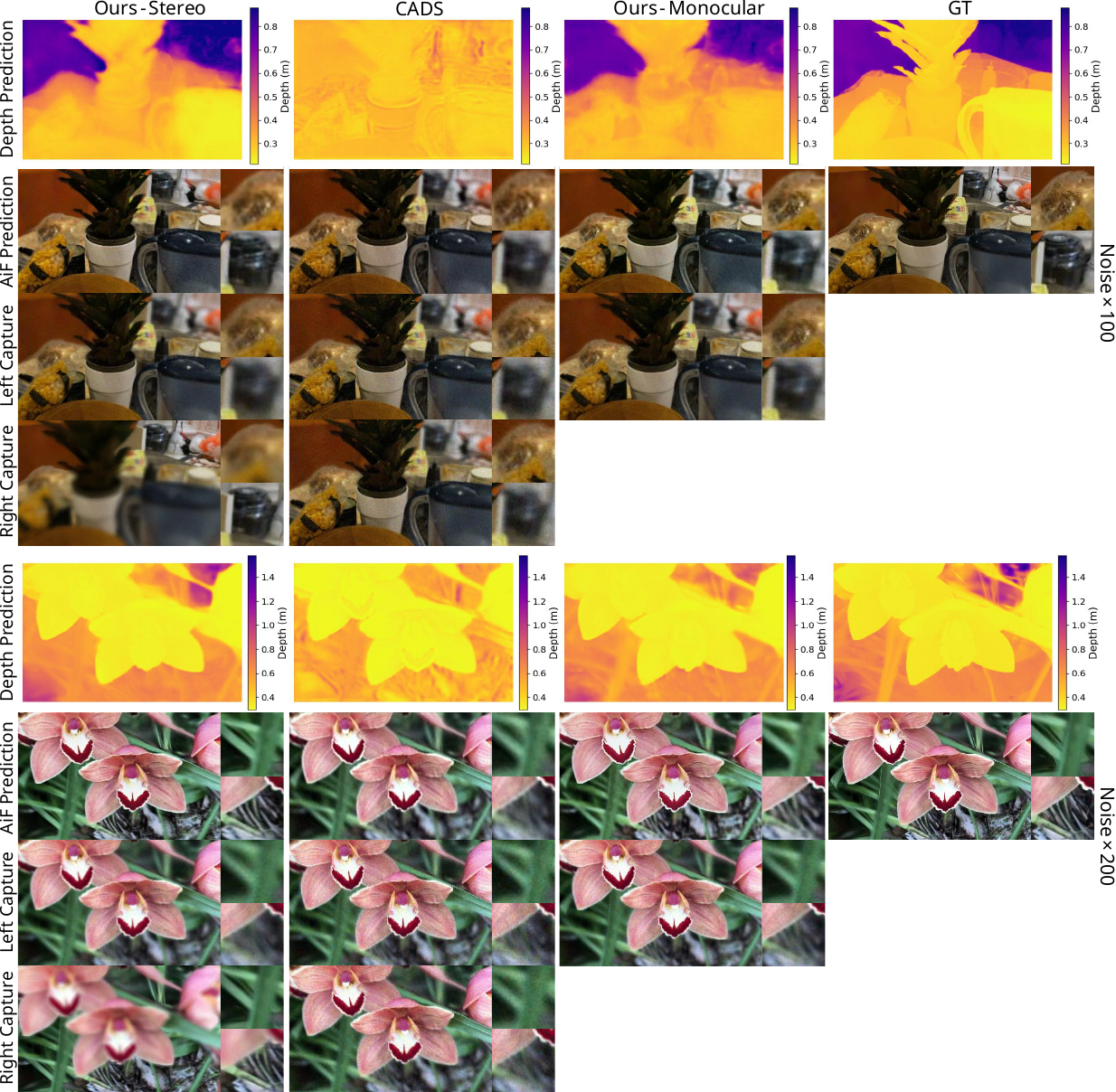}
        \caption{Qualitative comparison of the proposed stereo method with a coded aperture dual-pixel camera method (CADS~\cite{ghanekar2024passive}) and the monocular alternative. Our method achieves better AiF reconstruction by combining varying focus planes across the two views, and better depth estimation by exploiting disparity as an additional cue.
        }
\label{diff-fig:stereo_full}
\end{figure*}

\section{Additional Qualitative Results on Stereo Camera Design}
Fig.~\ref{diff-fig:stereo_full} presents full qualitative comparisons between our method and the baselines across both noise scenarios. Our stereo setup achieves higher AiF reconstruction quality by combining photometric cues from both views under strong defocus blur, with one view focused on the foreground and the other on the background. Its depth estimation is likewise improved by exploiting disparity as an additional cue. The baseline methods, in contrast, are limited to a single focal distance by their hardware, resulting in lower AiF reconstruction quality and depth estimation accuracy under strong defocus blur.

\begin{figure*}[!ht]
        \centering
        \includegraphics[width=\linewidth]{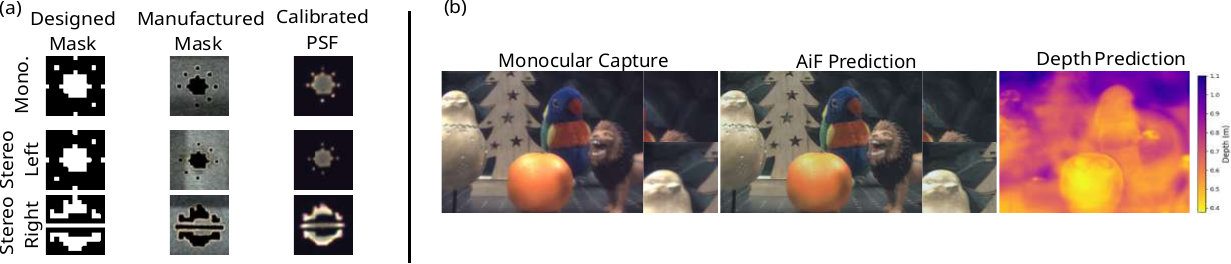}
        \caption{(a) The optimised coded aperture masks in both monocular and stereo experiments, the manufcatured masks, and their measured PSFs. (b) The captured coded aperture image, reconstructed AiF image, and estimated depth map for the monocular setup. The AiF and depth recovery on real hardware show plausible outcomes but with modest accuracy gap attributable to the sim-to-real domain shift and manufacturing tolerances.
        }
\label{diff-fig:real_mono}
\end{figure*}

\section{Additional Details on Real-World Experiments}
We validate our method with a real-world prototype of a coded aperture camera. As discussed in the main paper, we use a commercial FLIR Flea3 camera with a Fujinon DF6HA-1B lens due to manufacturing constraints. The lens has a fixed focal length of 6.15\,mm; we additionally maximise the entrance pupil diameter by setting the f-number to 1.2, which eases production of the coded aperture mask. We optimise the coded aperture mask and the lens's focus distance for both the monocular and stereo experiments, with the mask set to a 13$\times$13 grid. A relay lens (Edmund Optics \#46-007) then images the manufactured mask onto the entrance pupil of the main lens, placing the coded aperture at the pupil plane. The setup is shown in Figure 5 of the main paper.

The manufactured mask's PSF is measured by imaging a point light source at a known focus distance and source-to-camera distance. This measured PSF replaces its simulated counterpart in the image formation model, and we fine-tune the task network on the same training data for 1000 steps. Figure~\ref{diff-fig:real_mono}~(a) shows the optimised coded aperture mask for both monocular and stereo setups, the manufactured masks, and their measured PSFs.

Figure~\ref{diff-fig:real_mono}~(b) shows the captured coded aperture image, reconstructed AiF image, and estimated depth map for the monocular setup, while the corresponding stereo results appear in Figure 5 of the main paper. Our method reconstructs sharp AiF images and plausible depth maps under real optical conditions, though the depth estimation lags behind the simulated results, with visible errors near the scene's mid-plane. We attribute this shortfall to manufacturing tolerances and the broader gap between simulated and captured imagery. Comparing Figure~\ref{diff-fig:real_mono}~(b) with Figure 5 of the main paper, the stereo configuration yields higher depth accuracy than the monocular one, in line with our simulated results.

